\documentclass[runningheads]{llncs}

\usepackage{eccv}

\usepackage{eccvabbrv}

\usepackage{graphicx}
\usepackage{booktabs}
\usepackage{multirow}
\usepackage{makecell}
\usepackage{tabularx}
\usepackage{xurl}

\usepackage[accsupp]{axessibility}  % Improves PDF readability for those with disabilities.
\newcommand{\mbf}[1]{\mathbf{#1}}
\newcommand{\cut}[1]{}

\usepackage{subcaption}

\usepackage{hyperref}
\usepackage{orcidlink}

\begin{document}

\title{Probing the 3D Object-Level Understanding of Pre-Trained Detection Transformers} 

\titlerunning{Probing 3D Understanding of DETRs}

\author{Robin Kim\inst{1}\orcidlink{0009-0003-1070-4554} \and
Colin Samplawski \inst{2}\orcidlink{0000-0001-8486-1673} \and
Benjamin M. Marlin \inst{1}\orcidlink{0000-0002-2626-3410}}

\authorrunning{R. Kim et al.}

\institute{UMass Amherst, Amherst, MA, USA  \\ \email{\{seunghyunkim,marlin\}@cs.umass.edu} \and
SRI International, Menlo Park, CA, USA \\ \email{colin.samplawski@sri.com}}
% \email{colin.samplawski@sri.com}\\

\maketitle

\begin{abstract}
Detection transformer models, including DETR and its extensions, learn to output a set of object-level embeddings that can be simultaneously decoded into 2D bounding boxes and class distributions. In this paper, we investigate what pre-trained 2D detection transformers understand about the 3D properties of objects. Specifically, we investigate the extent to which properties including the depth of objects from the camera and the 3D location of objects relative to the camera can be recovered from object-level embeddings using linear and non-linear probes. Across a range of detection transformer models, our results show a surprisingly strong and previously unknown ability of 2D DETR models to represent  useful information about the 3D properties of objects, despite the complete lack of 3D supervision during model pre-training.


\end{abstract}
\section{Introduction}\label{sec:intro}

Detection transformer models including DETR and its extensions \cite{carion2020end, zhu2020deformable, meng2021conditional, zhao2024detrs, lv2024rt, chen2024lw} are examples of vision transformer models \cite{khan2022transformers} applied to the task of 2D object detection. Detection transformers take individual images as input, and produce as intermediate output a set of  embeddings corresponding to latent representations of potential objects in the image. Each query embedding is then decoded into a distribution over classes, and a bounding box in the image plane. Rather than relying on non-maximum suppression (NMS) techniques, the predicted class distribution for each candidate object can be thresholded to determine the final set of predicted objects. Like typical image detection models, the DETR model family is trained using image datasets annotated with classes and 2D bounding boxes \cite{lin2014microsoft} only.

The transformer-based encoder-decoder architecture expressed in the original DETR model \cite{carion2020end} has been extended to address several other object-oriented, image-based prediction tasks that go beyond predicting object class labels and 2D bounding boxes. Example tasks include instance segmentation \cite{maskformer_NEURIPS2021} and pose estimation \cite{tokenpose_2021_ICCV}, as well as monocular 3D object detection \cite{detr3d_crl22,zhang2023monodetr}. Models for such tasks typically use modified architectures with task-specific decoders or output heads, as well as suitably annotated training data.

In this paper, our primary interest is in the inference of 3D properties of objects from single images. However, unlike past work that has developed and trained models specifically for 3D prediction from images, we pose the question ``to what extent do 2D detection transformers encode information about the 3D properties of objects in their latent object embeddings?'' Specifically, we investigate the extent to which properties including the depth of objects from the camera and the 3D location of objects relative to the camera can be recovered from pre-trained object-level embeddings using linear and non-linear probes. 

Across a range of 2D detection transformer architectures, our results show a surprisingly strong and previously unknown ability of models to infer both depth and 3D location information. On the depth estimation task, we show that the pre-trained latent embeddings of 2D DETR models can be decoded using non-linear probes at a performance level that meets or exceeds that of a zero-shot application of the Depth Anything 2 visual foundation model \cite{yang2024depth}. On the task of 3D object location prediction, we show that pre-trained latent embeddings of 2D DETR models can be decoded using non-linear probes at an error level that is only fractions of a meter greater than the error of the MonoDETR model, which has an architecture and training approach designed explicitly for this task \cite{zhang2023monodetr}. Interestingly, our results further show that real-time and lightweight variants of DETR appear to have diminished ability to represent 3D object information compared to other DETR variants. 

The remainder of this paper is organized as follows. In Section \ref{sec:background}, we discuss related work and describe the models used in our evaluation. In Section \ref{sec:methods}, we describe our linear and non-linear probing methodology, and define the 3D properties that we investigate. In Section \ref{sec:experiments}, we present experiments and discuss results. 
\section{Background and Related Work}\label{sec:background}
In this section, we discuss related work and present background information on the models used in our evaluation. We begin with a description of the 2D detection task and the DETR family of models. We then discuss 3D prediction problems and related models. Finally, we discuss related probing studies.

\subsection{2D Detection Transformers}

\textbf{The 2D Detection Task.} In the standard 2D object detection task, a model is provided with a color input image $\mathbf{x} \in \mathbb{R}^{H \times W \times 3}$. The model issues a set of detections of the form $\mathbf{y}=\{(l_i, \mathbf{b}^2_i) \mid 1\leq i \leq m\}$ where $l_i\in\mathcal{C}$ is a discrete class label and $\mathbf{b}^2_i$ represents a 2D bounding box in terms of its center $(\mbf{b}^2_{x,i}, \mbf{b}^2_{y,i})$, width $\mbf{b}^2_{w,i}$, and height $\mbf{b}^2_{h,i}$.
The number of detections to issue $m$ must be predicted by the model. We will refer to the output space of a 2D detector (the set of sets of detections) as $\mathcal{Y}$. To train a standard 2D object detector, a model is provided with a dataset $\mathcal{D}_{tr}=\{(\mathbf{x}_n,\mathbf{y}_n)\mid 1\leq n \leq M\}$ where  $\mathbf{x}_n \in \mathbb{R}^{H \times W \times 3}$ is the $n^{th}$ training image and $\mathbf{y}_n = \{(l_{ni}, \mbf{b}^2_{ni})\mid 1\leq i \leq m_n\}$ is the set of ground truth bounding boxes and class labels for the objects present in image $n$.

\begin{figure}[t!]
    \centering
    \includegraphics[width=1\linewidth]{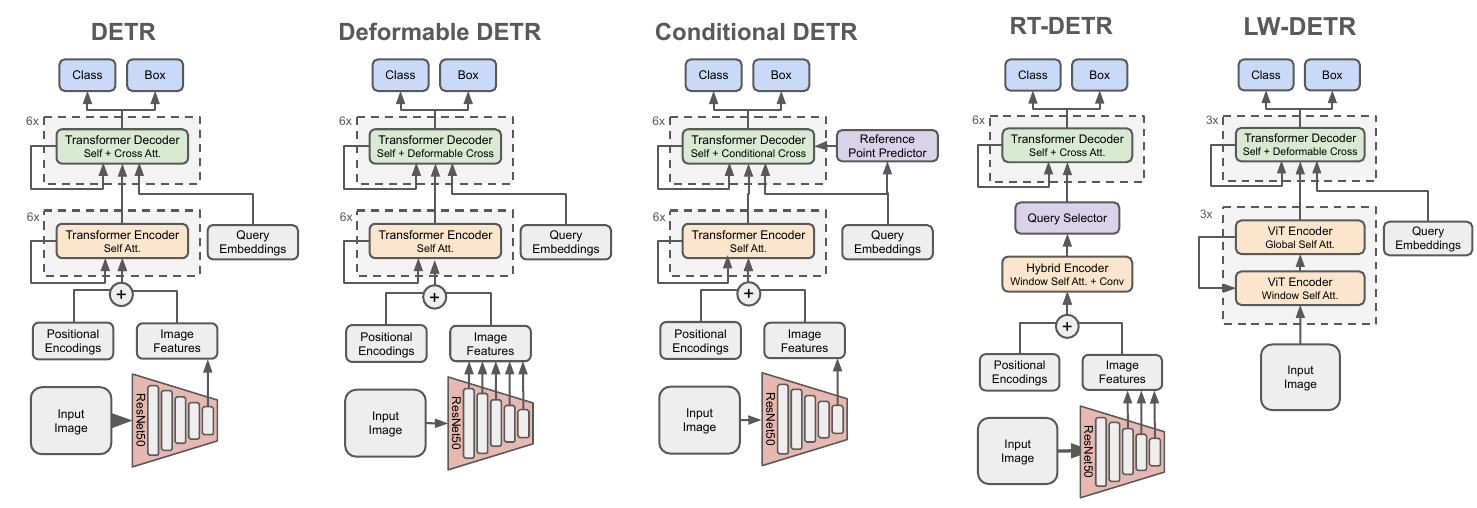}
    \caption{DETR model variants: DETR, Deformable DETR, Conditional DETR, Real Time DETR, Light Weight DETR.}
    \label{fig:detr_models}
\end{figure}

\textbf{The DETR Model.} Vision transformers (ViTs) \cite{khan2022transformers} have been applied to many computer vision tasks, including the 2D object detection task described above where they are referred to as detection transformers (DETRs) \cite{carion2020end, zhu2020deformable, meng2021conditional, zhao2024detrs, lv2024rt, chen2024lw}. DETRs generally follow the high-level architecture of the original DETR model proposed by \cite{carion2020end}. This architecture includes a deep backbone $f_{\theta}$ that extracts a pyramid of spatial features. In the original DETR model, the backbone is ResNet50 \cite{he2016deep}.

These features are then flattened into a sequence of visual tokens that are fed into a transformer encoder $e_{\phi}$ consisting of a series of self-attention layers that aggregate information across all visual tokens. The key innovation of the DETR family is the use of a transformer decoder $d_{\psi}$ that acts on a fixed set of $N$ query embeddings $q \in \mathbb{R}^{N \times d}$. These query embeddings are a set of free parameters that are fed into the decoder along with the visual tokens output by the encoder. The decoder consists of interleaved layers of self-attention on the queries, and cross-attention between the queries and the visual tokens. The original DETR model used transformer layers based on standard scaled dot product attention \cite{vaswani2017attention}.

The output of the decoder (still $\mathbb{R}^{N \times d}$) is then a set of updated queries which act as a set of object-level embeddings or tokens. Each embedding is then  regressed into object-level output representations using a collection of 
output heads. The original DETR model includes one output head $h^C_{\omega_C}$ that predicts a distribution over the object classes $\mathcal{C}$, and one output head $h^B_{\omega_B}$ that predicts the bounding box coordinates in $\mathbb{R}^4$. In the original DETR model, the class prediction head $h^C_{\omega_C}$ consists of a linear layer followed by a softmax mapping, while the bounding box prediction head is a small MLP. 

The DETR model architecture is sketched in Figure \ref{fig:detr_models} along with the additional model variants that we describe in the following sections.

\textbf{Deformable DETR.} A major shortcoming of the original DETR model is that it has poor training efficiency, requiring a large number of training epochs to achieve reasonable performance. This has been attributed to the standard scaled dot product attention layers used throughout the model, as they lack inductive bias for image-based data. To address this issue, the Deformable DETR model \cite{zhu2020deformable} instead uses a deformable attention layer. Here each position in the visual token sequence predicts a small number of other locations to attend to, across all feature scales. This allows for focused, global communication across the image plane, enabling faster  convergence during model training.  

\textbf{Conditional DETR.} Following a similar logic to Deformable DETR, the Conditional DETR model \cite{meng2021conditional} maintains the dense self-attention of the original DETR model, but adds an additional reference point prediction inside the decoder. Here a small auxiliary network predicts a point in a 2D normalized grid from each initial query embedding. These 2D reference points are then converted into sinusoidal positional embeddings. The positional embeddings are appended onto the query embeddings during the cross-attention operations inside each decoder layer, enabling more efficient localization within the image, and therefore faster training.

\textbf{RT-DETR.} An additional significant issue with the original DETR model is that it is significantly slower at prediction time than detection models using traditional CNN backbones, and explicit non-maximum suppression such as the YOLO family of detectors \cite{redmon2016you}. This led to several modified DETR architectures that aim to increase detection speed, including the Real-Time DETR (RT-DETR) model \cite{zhao2024detrs,lv2024rt}. The primary insight in the RT-DETR model architecture is that the simultaneous intra-scale and inter-scale attention used in Deformable DETR is not needed in 2D detection tasks. The model architecture instead uses a CNN to fuse information across scales, and attention to fuse information within a restricted set of scales. In addition, the approach uses a query selection scheme to initialize queries based on initial classification confidence estimates. The approach has been shown to be several times faster than deformable DETR variants, and on par with the run time of recent YOLO model variants \cite{zhao2024detrs}.

\textbf{LW-DETR.} Light Weight DETR (LW-DETR) is a DETR variant that, like RT-DETR, focuses on prediction speed. While RT-DETR retains a CNN backbone, LW-DETR instead uses a ViT backbone. To deal with the quadratic computation of the ViT backbone, LW-DETR interleaves global attention with windowed attention. The LW-DETR architecture uses a transformer decoder with deformable cross attention. Unlike other DETR architectures, the decoder uses only three layers to further reduce latency.

\subsection{3D Vision Transformers}

\textbf{3D Prediction Tasks.} Given an input image $\mathbf{x}$, there are multiple 3D prediction tasks of interest. The closest task to the 2D detection task is the 3D detection task, which consists of predicting the class label along with a 3D bounding box for each object in the image. Formally, the output of a 3D detector is given by $\mathbf{y}=\{(l_i,\mbf{b}^3_i,\phi_i) \mid 1\leq i \leq m\}$  where $l_i\in\mathcal{C}$ is a discrete class label, $\mbf{b}^3_i$ specifies the center and extent of a 3D bounding box $\mbf{b}^3_i = (\mbf{b}^3_{x,i}, \mbf{b}^3_{y,i}, \mbf{b}^3_{z,i}, \mbf{b}^3_{l,i}, \mbf{b}^3_{w,i}, \mbf{b}^3_{h,i})$, and $\phi_i=(\phi^{y}_i,\phi^{p}_i,\phi^r_i)$ represents the yaw, pitch and roll angles of the bounding box in the camera's local 3D coordinate system. We note that models developed for specific applications may only include a subset of these outputs.

Another 3D prediction task of interest is inferring a dense depth map given an input image $\mathbf{x}\in\mathbb{R}^{H \times W \times 3}$. The output of this prediction task is a single matrix $\mathbf{y}\in\mathbb{R}^{H \times W}$ giving the depth of the surface represented by each pixel location $(i,j)$ in $\mathbf{x}$. While the depth map prediction task is not an object-oriented task, the output of the depth map prediction task can be combined with the output of a 2D detection model to extract information about the depth of detected objects from the camera. 

\textbf{DETR 3D.} The DETR 3D model \cite{detr3d_crl22} outputs per-object 3D bounding boxes following a generalization of the 3D detection task described above. Specifically, the input consists of multiple images of the same scene. The model architecture leverages components of the general DETR framework including a ResNet backbone, and transformer self-attention layers. However, the architecture is decoder-only following the backbone, and the main innovation is the use of camera intrinsic and extrinsic information to project queries and features back and forth between 2D and 3D representations. The model is trained using 3D box ground truth.

\textbf{MonoDETR.}  The MonoDETR model addresses the 3D prediction task using single image inputs as described above \cite{zhang2023monodetr}. The MonoDETR architecture uses multi-scale features from a ResNet50 backbone, followed by distinct encoder pathways for RGB image features and depth features. The outputs of the two encoder pathways are fused in the decoder. Like DETR, the decoder uses learned query embeddings, but alternates between attention computations based on image embeddings and depth embeddings. The model is trained on a combination of sparse object-level 2D depth data, and 3D box ground truth. Training and evaluation focus on the KITTI dataset \cite{zhang2023monodetr}.

\textbf{Depth Anything.} The Depth Anything model family addresses the task of predicting dense depth maps from individual images \cite{yang2024depth}. The model combines a DINOv2 (self-DIstillation with NO labels) vision transformer encoder \cite{oquab2024dinov2} with a DPT (Dense Prediction Transformer) decoder \cite{ranftl2021vision}. The primary innovation in the model is the training procedure, which uses large-scale synthetic image/depth data to train a base model, followed by the use of large-scale real data to drive a distillation process that enables producing accurate lighter-weight models.

\subsection{Probing of Deep Models}
Due to the poor interpretability of modern deep networks, probing techniques have emerged as a method to infer what concepts are encoded by the latent representation of a pre-trained model \cite{alain2016understanding}. Probing techniques typically train a simple linear or MLP model to predict a property of interest using frozen, pre-trained intermediate features as input. The lower the prediction error of the trained probe model, the greater the degree to which the frozen features encode information about the property of interest. In recent years, these techniques have become a particularly popular choice for interpreting large language models \cite{marks2023geometry,azaria2023internal,park2024linear}. Naturally, these techniques have also been applied to recent vision models \cite{gairola2025probe,li2023evaluating,caron2021emerging}, including (CNN-based) object detectors \cite{wagner2024forgetting,mikriukov2023evaluating}.

More specifically, prior work has investigated whether recent vision foundation models encode 3D and physical scene understanding. For example, \cite{zhan2024general,el2024probing} studied 3D awareness of pre-trained vision foundation models including \mbox{DINOv2}\cite{oquab2024dinov2} and Stable Diffusion\cite{rombach2022high}. They both conclude that DINOv2 and Stable Diffusion encode some degree of 3D knowledge, even though they are trained without 3D supervision. The DepthCues benchmark of \cite{danier2025depthcues} provides a number of tasks which test a probe's ability to understand 3D information based on a single image. These works show that pre-trained visual representations contain significant geometric, physical, and 3D-specific knowledge.

The focus of our work is different from these prior works. They primarily probe generic backbone features, dense feature maps, or pooled region/scene representations. Even when object-level tasks are considered, the unit of analysis is not the detector's query embeddings. Even prior work that explicitly probes object detectors only considers CNN-based detectors \cite{wagner2024forgetting,mikriukov2023evaluating}. To the best of our knowledge, no prior work has considered probing the query embeddings output by transformer-based detectors. This paper addresses this gap by focusing on the object-level 3D understanding of the DETR family of 2D object detection models.

% \newpage

\section{Methods}\label{sec:methods}
In this section, we describe the models, 3D properties, and datasets used as the basis for our experiments.

\subsection{Models, Query Embeddings and 3D Properties}
\label{sec:query-emb}

Our goal in this work is to investigate the extent to which pre-trained 2D detection transformers learn latent representations of 3D properties of objects. We focus on the five DETR model variants described in Section \ref{sec:background}: DETR, Deformable DETR, Conditional DETR, RT-DETR v2, and LW-DETR. These models provide a representative sample of 2D detection transformer architectures, exhibiting different choices for the backbone model, the use of multi-scale features, the use of different attention mechanisms, and the use of different encoder and decoder depths. We summarize properties of the models we use in Table \ref{tab:models}.

As described in Section \ref{sec:background}, all of these models produce a set of embeddings $\{\mathbf{q}_i \in \mathbb{R}^D\}_{i=1}^{N}$ of dimension $D$ for each of $N$ candidate objects as an intermediate representation. The dimension $D=256$ for all models considered, except for LW-DETR where  $D=384$. We operationalize our goal of determining the extent to which pre-trained 2D detection transformers learn latent representations of 3D properties of objects in terms of the extent to which 3D properties of objects can be predicted from object embeddings $\mathbf{q}_i$. We address this question for select 3D properties using the probing approach described in the next section. In this work, we focus on two fundamental 3D properties of objects, their depth from the camera and their location in 3D space. 

\begin{table}[t]
  \centering
  \caption{Overview of models used in this study. DETR models in the top section of the table are trained without 3D supervision. Depth Anything 2 and MonoDETR are trained using different forms of 3D supervision and serve as baselines for 3D tasks.}
  \label{tab:models}
  \small
  \setlength{\tabcolsep}{5pt}
  \begin{tabular}{l l l l}
  \toprule
  \textbf{Model} & \textbf{Backbone} & \textbf{Supervision} & \textbf{Dataset} \\
  \midrule
  DETR~\cite{carion2020end}             & ResNet  & Supervised & ImageNet + COCO \\
  LW-DETR~\cite{chen2024lw}           & ViT      & SSL + Supervised & Object365 + COCO \\
  RT-DETR v2~\cite{lv2024rt}            & ResNet  & Supervised & ImageNet + COCO \\
  Conditional DETR~\cite{meng2021conditional} & ResNet  & Supervised & ImageNet + COCO \\
  Deformable DETR~\cite{zhu2020deformable} & ResNet  & Supervised & ImageNet + COCO \\
  \midrule
  \midrule
  Depth Anything 2~\cite{yang2024depth} & ViT & SSL + Supervised & Synthetic + Real \\
  MonoDETR~\cite{zhang2023monodetr} & ResNet  & Supervised & ImageNet + KITTI3D \\
  \bottomrule
  \end{tabular}
  \end{table}

\subsection{Probing Methodology}
\label{sec:probing}

In order to assess the extent to which 3D information is encoded in the object embeddings of 2D DETR models, we adopt a probing approach that has been widely used in the analysis of latent representations~\cite{alain2016understanding,belinkov2022probing}. Given a collection of $K$ 3D object-level properties $P$ that we would like to jointly assess, we begin by constructing a training dataset $\mathcal{D}^{P}_{tr} = \{(\mathbf{q}_n,\mathbf{p}_n) \mid 1\leq n\leq M_{tr}\}$ where each $\mbf{q}_n \in \mathbb{R}^D$ is an  object embedding and each $\mbf{p}_n \in \mathcal{P}$ is the corresponding vector of property values. The set $\mathcal{P}$ represents the space of possible property vectors. 

Next, we select a probing model $f_{\theta}: \mathbb{R}^D \rightarrow \mathcal{P}$. In this work, we consider linear probe models $f^{lin}_{\theta}$, and two-layer fully-connected neural network (MLP) probe models using ReLU activations $f^{mlp}_{\theta}$. We use the training dataset $\mathcal{D}^{P}_{tr}$ to learn the parameters $\theta$ of the probe model $f_{\theta}$, and then use a held out test dataset of object embeddings and property vectors $\mathcal{D}^{P}_{te} = \{(\mathbf{q}_n,\mathbf{p}_n) \mid 1\leq n\leq M_{te}\}$ to assess the performance of the learned probe. 

For all experiments, the detector weights are frozen and only the probe is trained. The linear probe consists of a single affine map from the detector embedding to the target space. The non-linear probe is a two-layer MLP with ReLU activations and hidden dimension $256$. Unless otherwise noted, we use a batch size of $512$, learning rate $10^{-3}$, and mean squared error as the optimization objective for regression tasks. For the MLP depth experiments, we use $1000$ epochs with a $20$-epoch warm up followed by cosine decay. We use Adam as the optimizer. 

We report predictive performance on the test set in terms of mean absolute error $\mathrm{MAE}=\frac{1}{N}\sum_i|\hat y_i-y_i|$, threshold accuracy $\delta_1=\frac{1}{N}\sum_i \mathbb{I}\!\left[\max(\hat y_i/y_i,\, y_i/\hat y_i)<1.25\right]$, and absolute relative error $\mathrm{AbsRel}=\frac{1}{N}\sum_i\frac{|\hat y_i-y_i|}{y_i}$ for each property.   

The next question is how to construct the necessary datasets for properties of interest given that we need to relate the object embeddings predicted by multiple pre-trained detectors to properties of the predicted objects. We use different approaches for different probe tasks based on the affordances of existing datasets, as described in the next section.

\subsection{Probe Dataset Construction}
\label{sec:probe-data}

Our general approach to constructing the datasets $\mathcal{D}^{P}_{tr}$ and $\mathcal{D}^{P}_{te}$ needed to train and test probe models for different 3D properties leverages base training and test sets  $\mathcal{D}^B_{tr}=\{(\mathbf{x}_n,\mathbf{y}_n)\mid 1\leq n \leq M_{tr}\}$ and $\mathcal{D}^B_{te}=\{(\mathbf{x}_n,\mathbf{y}_n)\mid 1\leq n \leq M_{te}\}$ where each $\mathbf{x}_n$ is an image and each $\mathbf{y}_n$ is a task-specific annotation structure. For each image $\mathbf{x}_n$ in $\mathcal{D}^B_{tr}$, we run a pre-trained 2D DETR model to obtain the object embeddings $\mbf{q}_{ni}$, class label probability distributions $\mbf{l}_{ni}$ and 2D bounding boxes $\mbf{b}^2_{ni}$. 
We consider an embedding $\mbf{q}_{ni}$ to predict the existence of an object if the predicted probability that the object belongs to the set of foreground classes as specified by $\mbf{l}_{ni}$ exceeds a threshold $\tau$. 

For each query embedding $\mbf{q}_{ni}$ that passes the objectness threshold, we extract the needed property values $\mbf{p}_{ni}$ from the annotation structure $\mbf{y}_n$ for training image $n$. This allows us to form the training pairs $(\mbf{q}_{ni},\mbf{p}_{ni})$ needed to construct $\mathcal{D}^{P}_{tr}$. We use the same process on the base test dataset $\mathcal{D}^B_{te}$ to form the dataset of test instances $\mathcal{D}^{P}_{te}$. We provide details for specific property sets below.

\textbf{Monocular Depth Estimation.} For the case of monocular depth estimation, we define the 3D property of interest to be $D$, the depth from the camera at the center of the predicted bounding box for each predicted object. To build the datasets $\mathcal{D}^{D}_{tr}$ and 
$\mathcal{D}^{D}_{te}$, we assume access to base datasets where each $\mathbf{x}_n$ is an image and each $\mathbf{y}_n$ is a dense depth map (either ground truth, or predicted by a dense depth model). For each query embedding $\mbf{q}_{ni}$ that passes the objectness threshold, we extract the depth value $d_{ni}$ at the predicted bounding box center location $(\mbf{b}^2_{x,ni},\mbf{b}^2_{y,ni})$ from the dense depth map $\mbf{y}_n$. This allows us to form the training pairs $(\mbf{q}_{ni},d_{ni})$ needed to construct $\mathcal{D}^{D}_{tr}$. We use the same process on the base test dataset $\mathcal{D}^B_{te}$ to form the dataset of test instances $\mathcal{D}^{D}_{te}$.

\textbf{3D Location Prediction.} The second task that we consider is 3D location prediction. For this task, the property of interest is the location of the 3D center $C$ of an object in the coordinate system of the camera. To build the datasets $\mathcal{D}^{C}_{tr}$ and 
$\mathcal{D}^{C}_{te}$, we assume access to base datasets where each $\mathbf{x}_n$ is an image and each $\mathbf{y}_n$ is a set of annotations including 2D and 3D bounding boxes for each object present in each image. We extract the set of object embeddings $\mbf{q}_{ni}$ that passes the objectness threshold for each image. We then compute the IoU of each predicted 2D bounding box $\mbf{b}^2_{ni}$ with each ground truth bounding box $\mbf{b}^2_{nj}$. We rank the pairs and perform a greedy matching. Predicted boxes that do not match any ground truth box are discarded. For each object embedding $\mbf{q}_{ni}$ that is matched with a ground truth object $j$ in image $n$, we extract the 3D bounding box center $(\mbf{b}^3_{x,nj},\mbf{b}^3_{y,nj}, \mbf{b}^3_{z,nj})$ as its property vector.

\textbf{Dataset Alignment.} Since the dataset construction method described above is based on filtering detections produced by a given pre-trained DETR model, the approach will result in training and test datasets for different sets of predicted objects. To facilitate comparisons between DETR model variants in some experiments, we align the probing datasets across models after they are constructed using a second level filtering process. 

Specifically, we select one DETR model to use as an anchor model. For each image $n$, we compute the IoU between each 2D bounding box $\mbf{b}^2_{ni}$ predicted by the anchor model and each 2D bounding box $\mbf{b}^2_{nj}$ predicted by an alternate model. We threshold the matches at an IoU of $0.5$, rank the surviving pairs, and greedily match them. Lastly, for each predicted object $i$ produced by the anchor model for each image $n$, we require that it be matched to some prediction $j$ produced by each of the alternate models on the same image. For each model, we discard all instances from its training and test sets $\mathcal{D}^P_{tr}$ and $\mathcal{D}^P_{te}$ that do not satisfy this criterion, resulting in new filtered datasets $\mathcal{F}^P_{tr}$ and $\mathcal{F}^P_{te}$. The filtered datasets include an intersection set of objects that are detected simultaneously by all models.

\section{Experiments}\label{sec:experiments}
In this section, we describe experiments and results assessing the extent to which pre-trained 2D DETR models contain latent representations of 3D properties of objects. As described in Section \ref{sec:methods}, we focus on monocular depth estimation and 3D object center prediction as probing tasks. The code and experiment configurations are available at \url{https://github.com/reml-lab/DETRProbe3D/}.

\subsection{Monocular Depth Estimation}

\textbf{Experimental Protocol.} We evaluate object-level monocular depth understanding using probing datasets constructed from two widely used monocular depth estimation base datasets: Virtual KITTI\,2~\cite{cabon2020virtual}, and NYU Depth v2 (NYUv2)~\cite{silberman2012indoor}. Virtual KITTI\,2 provides synthetic outdoor driving scenes of resolution \(1242\times 375\) with dense metric depth, and multiple weather and viewpoint variants. NYUv2 consists of \(640\times 480\) real indoor RGBD pairs, allowing us to test depth understanding for indoor scenes. We apply the monocular depth estimation probe dataset construction approach described in Section \ref{sec:methods} with an objectness threshold of $\tau=0.5$. For Virtual KITTI\,2, we keep only detections mapped to the \texttt{car} class. For NYUv2, we keep all detected classes. We filter out instances with depth values outside of 0.5--100\,m for Virtual KITTI\,2 and outside of 0.1--10\,m for NYUv2 to minimize noise in true depth values. 

For this experiment, we apply the dataset alignment process described in  Section \ref{sec:methods} using DETR as the anchor model. The aligned dataset sizes for Virtual KITTI\,2 are (\(N_\text{train}=3012\), \(N_\text{test}=754\)) and for NYUv2 (\(N_\text{train}=4600\), \(N_\text{test}=1150\)). We additionally run Depth Anything 2 on the same images and sample its depth predictions at the anchor model's box centers, producing a center aligned depth baseline. Similarly, we apply MonoDETR both zero-shot, and with a learned MLP output head matching the architecture of the 2D DETR model probes. We use model checkpoints from the Hugging Face library \cite{wolf2020transformers}. All detector weights are loaded directly from public Hugging Face checkpoints using \texttt{from\_pretrained} with model-specific processor classes. 
%Please refer to Appendix for the specific checkpoints.
%Please refer to Appendix for more details on building the dataset.

\begin{table}[t]
    \centering
    \caption{Monocular depth estimation results. Best values obtained by probing 2D detection models are shown in \textbf{bold}. Best values obtained from 3D models are \underline{underlined}. These results show that DETR family latent representations contain significant object-level depth information.}
    \label{tab:depth}
    \scriptsize
    \setlength{\tabcolsep}{4pt}
    \resizebox{\linewidth}{!}{%
    \begin{tabular}{ll ccc ccc}
    \toprule
    & & \multicolumn{3}{c}{\textbf{Virtual KITTI\,2 (Outdoor)}} & \multicolumn{3}{c}{\textbf{NYUv2 (Indoor)}} \\
    \cmidrule(lr){3-5} \cmidrule(lr){6-8}
    \textbf{Model} & \textbf{Probe} & MAE\,$\downarrow$ & AbsRel (\%)\,$\downarrow$ & $\delta_1$ (\%)\,$\uparrow$ & MAE\,$\downarrow$ & AbsRel (\%)\,$\downarrow$ & $\delta_1$ (\%)\,$\uparrow$ \\
    \midrule
    DETR & Linear & 1.01 & 8.29 & 93.10 & 0.58 & 24.30 & 57.22 \\
    Conditional-DETR & Linear & 1.13 & 9.07 & 92.57 & 0.53 & 22.52 & 60.96 \\
    Deformable-DETR & Linear & 1.06 & 8.34 & 91.78 & 0.56 & 23.45 & 59.91 \\
    LW-DETR & Linear & 1.26 & 9.84 & 91.38 & 0.59 & 25.30 & 58.70 \\
    RT-DETR v2 & Linear & 1.56 & 12.49 & 86.21 & 0.64 & 26.80 & 55.22 \\
    \midrule
    DETR & MLP & \textbf{0.54} & \textbf{3.62} & 98.81 & {0.51} & {20.20} & {65.48} \\
    Conditional-DETR & MLP & 0.69 & 4.39 & 98.67 & \textbf{0.50} & \textbf{19.99} & \textbf{66.78} \\
    Deformable-DETR & MLP & 0.65 & 4.05 & \textbf{98.94} & 0.54 & 21.49 & 63.65 \\
    LW-DETR & MLP & 0.88 & 5.63 & 98.01 & 0.55 & 22.65 & 62.70 \\
    RT-DETR v2 & MLP & 1.05 & 6.73 & 97.21 & 0.56 & 22.70 & 60.43 \\
    \midrule
    \midrule
    % DETR Bbox probe (Linear)  & 4.27 & 32.67 & 44.69 & 0.87 & 37.61 & 41.74 \\
    % DETR Bbox probe (MLP) & 1.02 & 7.09 & 95.76 & 0.77 & 32.98 & 51.83 \\
    Depth Anything 2 (Zero-shot)& - & 1.49 & 9.13 & 97.88 & 0.61 & 24.21 & 59.57 \\
    MonoDETR (Zero-shot) & - & 3.83 & 39.82 & 64.72 & 10.59 & 497.72 & 0.00 \\
    \midrule
    MonoDETR (MLP head) & - & \underline{0.47} & \underline{3.14} & {99.42} & 0.75 & 21.49 & 62.5 \\
    Depth Anything 2~(MLP head) & - & 0.50 & 3.30 & \underline{99.47} & \underline{0.36} & \underline{13.36} & \underline{84.00} \\
    \bottomrule
    \end{tabular}
    }
    \end{table}

\textbf{Results} Table~\ref{tab:depth} presents the monocular depth estimation results when applying the probe to the last decoder layer of each DETR model. Overall, the DETR-family latent representations produce competitive depth predictions on both datasets using both linear and MLP probes. We can clearly see that MLP probes outperform linear probes, indicating that the relationship between latent representations and depth estimates is non-linear. We note that among the DETR variants, the real-time models (RT-DETR v2 and LW-DETR) show weaker performance on Virtual KITTI\,2 compared to the other three DETR variants. A plausible explanation is that these architectures reduce layers and capacity to meet latency constraints, which may limit the richness of depth cues encoded in the object representations. Despite differences in backbone and attention structures, DETR, Deformable DETR, and Conditional DETR exhibit fairly similar  performance. 

Further, we see that when Depth Anything 2 and MonoDETR are applied zero-shot on these datasets, they are both out-performed in terms of MAE by all 2D DETR models combined with the MLP probe. While this is surprising as both Depth Anything 2 and MonoDETR are trained with 3D supervision, we note that both models experience domain shift when applied to our specific tasks. To address this, we learn modified output heads for both models with the rest of the parameters frozen to enable fine-tuning to our tasks. For Depth Anything 2, we spatially rescale the patch of output depth values inside the predicted bounding box of each object to a constant size, and learn an MLP output head on this representation. For MonoDETR, we add an MLP output head on top of the last decoder layer representation. Both MLPs have an architecture matching that of the MLP probes used with the 2D DETR models. The results are shown in the bottom two rows of Table~\ref{tab:depth}. We can see that on the Virtual KITTI 2 dataset, both models improve significantly using output heads learned for this task. On the NYUv2 dataset, Depth Anything 2 improves significantly, while MonoDETR is still out-performed by some of the DETR models with MLP probes. The continued weak performance of MonoDETR on NYUv2 is likely attributable to the objects in NYUv2 being out of domain relative to the dataset used to train MonoDETR. Overall, despite the improved performance of Depth Anything 2 and MonoDETR when fine-tuned for the current tasks, the depth estimation performance of the DETR models combined with MLP probes remains surprisingly and objectively strong given the total lack of 3D supervision during latent representation learning.

\begin{figure}[t]
    \centering
    \begin{subfigure}[t]{0.48\linewidth}
        \centering
        \includegraphics[width=\linewidth]{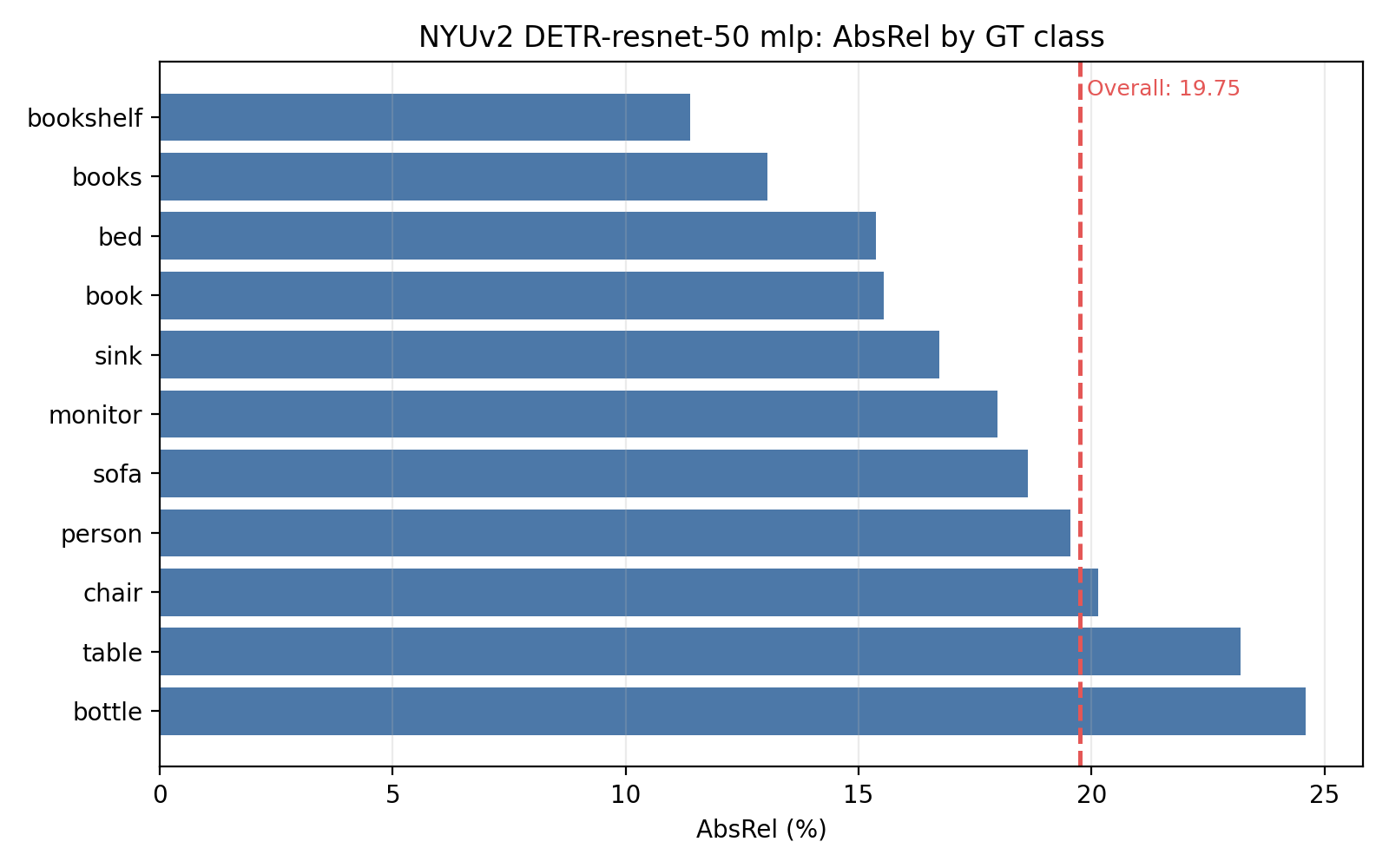}
    \end{subfigure}
    \hfill
    \begin{subfigure}[t]{0.48\linewidth}
        \centering
        \includegraphics[width=\linewidth]{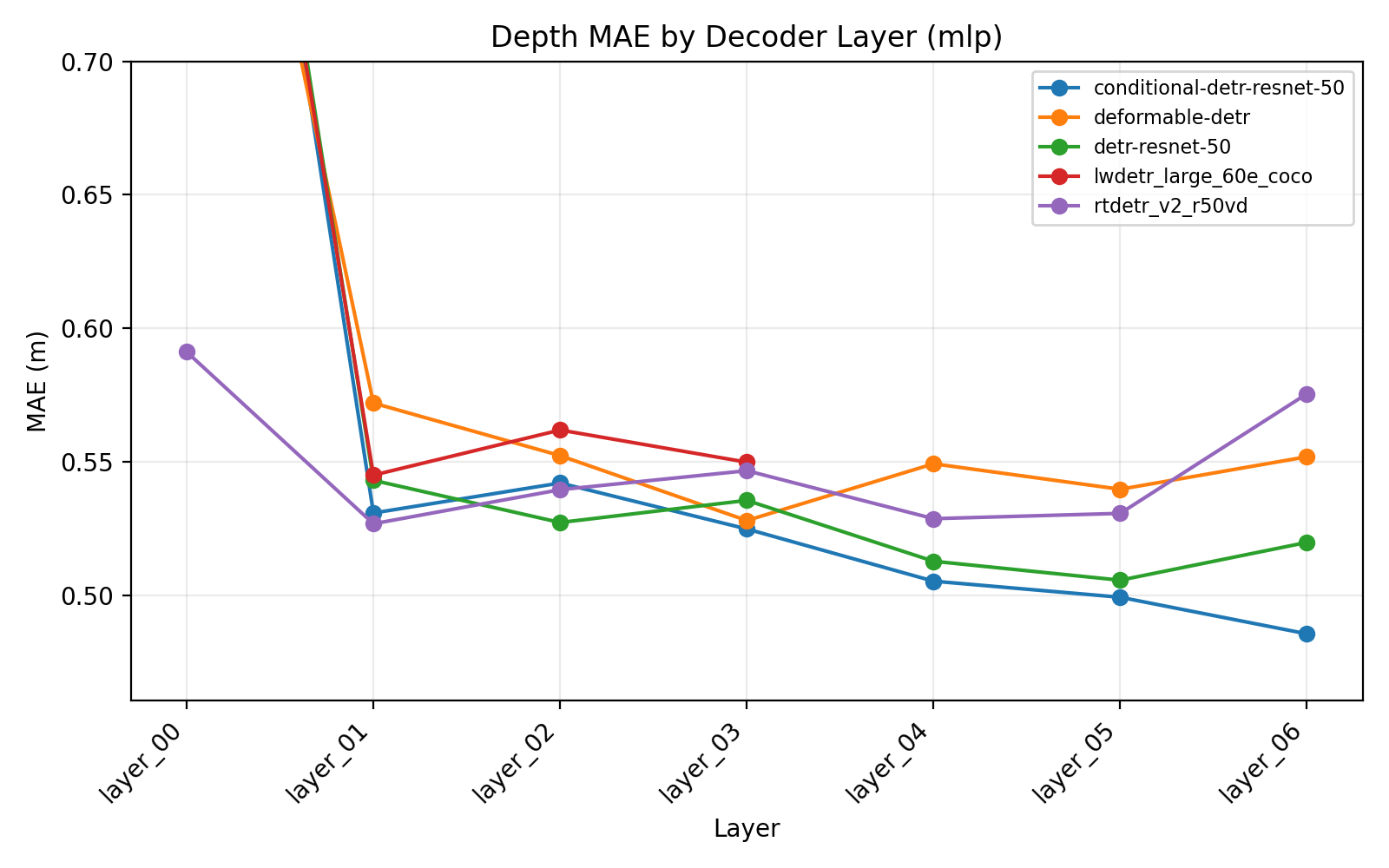}
    \end{subfigure}
    \caption{Further analysis on NYUv2 dataset. (Left) Classwise absolute relative depth prediction error for DETR. (Right) Layerwise depth mean absolute error for all DETR models.}
    \label{fig:more-nyuv2-analysis}
\end{figure}

Next, we consider the question of how depth estimation performance varies by object class. While our evaluation based on the Virtual KITTI\,2 dataset is restricted to a single class, the NYUv2 dataset includes multiple classes. In Figure~\ref{fig:more-nyuv2-analysis}(Left), we show how depth estimation error varies across object classes within the NYUv2 dataset for the original DETR model. These results indicate differences in absolute relative error for different classes. 

Finally, we investigate the question of whether probing each DETR model at the last decoder layer is optimal. We report depth estimation MAE on the NYUv2 dataset based on MLP probing at different decoder layers in Figure~\ref{fig:more-nyuv2-analysis} (Right). These results show that the optimal decoder layer to use for depth probing varies by model type, with some models exhibiting modest improvements in depth MAE at lower layers. Since the RT-DETR decoder initializes the query using tokens from the encoder, even probes on the earliest decoder layer show competitive performance.

\subsection{3D Location Prediction}

\textbf{Experimental Protocol.}
We evaluate 3D object location understanding using a probing dataset constructed
from the widely-adopted KITTI~\cite{geiger2012we} benchmark, following the data preprocessing protocol from \cite{zhang2023monodetr}. We follow the 3D location prediction dataset construction approach described in Section~\ref{sec:methods} with an objectness threshold of $\tau =0.5$. For this experiment, we report results for MLP probes only as linear probes again perform poorly. We do not align the datasets for each model to each other as they are already aligned against ground truth objects. We report the MAE metric for each dimension of the object center separately, as well as the MAE between the ground truth 3D object center and the predicted center. We note that we use the definition of the 3D bounding box center used in the KITTI dataset, which corresponds to the bottom center of each 3D box.

\begin{table}[t]
  \centering
  \caption{3D location prediction results. Error is reported for each component of the object center separately, as well as for the
  full vector of object center coordinates. Center AbsRel is reported as a percentage. Best values obtained by probing 2D detection models are shown in \textbf{bold}. Best values obtained from 3D models are \underline{underlined}.}
  \label{tab:kitti-center-vs-monodetr}
  \small
  \resizebox{\linewidth}{!}{
  \begin{tabular}{l c c c c c}
  \toprule
  \textbf{Model} & xMAE (m) $\downarrow$ & yMAE (m) $\downarrow$ & zMAE (m) $\downarrow$ & Center MAE (m) $\downarrow$ & Center AbsRel
  (\%) $\downarrow$ \\
  \midrule
  DETR & \textbf{0.33} & \textbf{0.12} & \textbf{1.03} & \textbf{0.50} & \textbf{2.40} \\
  Conditional-DETR & 0.68 & 0.16 & 1.09 & 0.64 & 3.01 \\
  Deformable-DETR & 0.66 & 0.16 & 1.04 & 0.62 & 2.90 \\
  RT-DETR v2 & 2.27 & 0.18 & 1.39 & 1.28 & 6.49 \\
  LW-DETR & 2.28 & 0.17 & 1.20 & 1.22 & 6.01 \\

  \midrule
  \midrule
  MonoDETR & \underline{0.19} & \underline{0.06} & \underline{0.64} & \underline{0.30} & \underline{1.41} \\
  \bottomrule
  \end{tabular}
  }
  \end{table}

\textbf{Results} Table~\ref{tab:kitti-center-vs-monodetr} reports 3D object location results for all of the 2D DETR models as well as for MonoDETR. Among the 2D DETR family models, DETR achieves the best performance. Deformable-DETR and Conditional-DETR have similar performance, while the real-time variants are again noticeably worse. While MonoDETR achieves the best performance on this task due to learning all model parameters using full 3D bounding box ground truth on this task, the results for the best performing DETR models with MLP probes are again surprisingly strong. The overall MAE for DETR is 0.5m compared to MonoDETR's 0.3m. We emphasize that the $0.2$m MAE difference between the best DETR MLP probe and MonoDETR on KITTI 3D is quite small. The median distance to the objects in our test dataset is $25.3$m. As we can see in Table~\ref{tab:kitti-center-vs-monodetr}, the gap in absolute relative error between the methods is about 1\%. This result again suggests that DETR's latent object representations encode significant 3D information despite the total lack of 3D supervision during latent representation learning.

\subsection{Further Analysis}

\begin{figure}[t]
\centering
\begin{minipage}{1.00\linewidth}
\centering
\includegraphics[width=\linewidth]{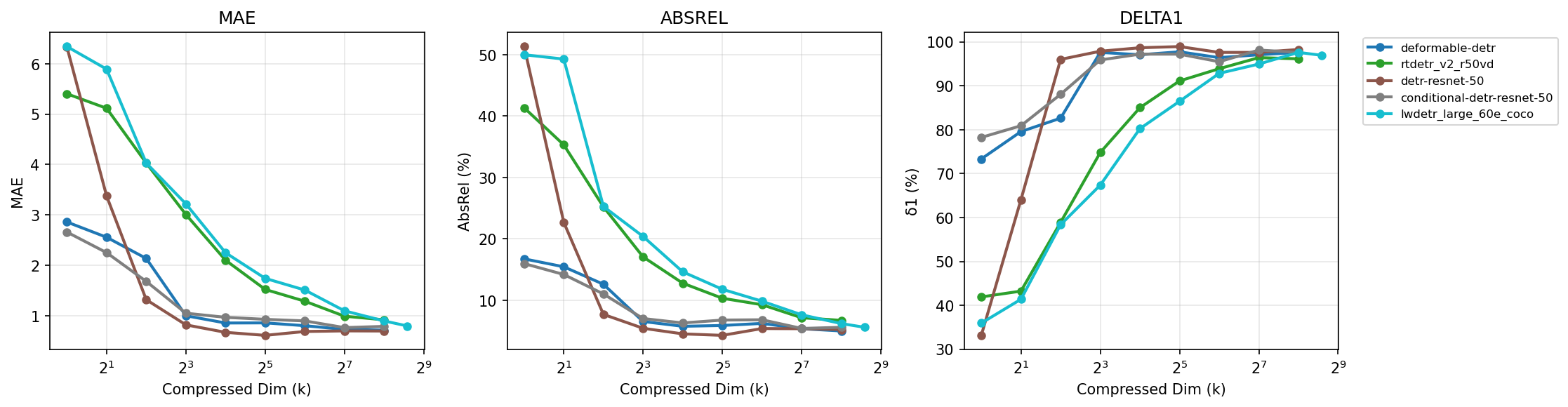}
\end{minipage}

\begin{minipage}{1.00\linewidth}
\centering
\includegraphics[width=\linewidth]{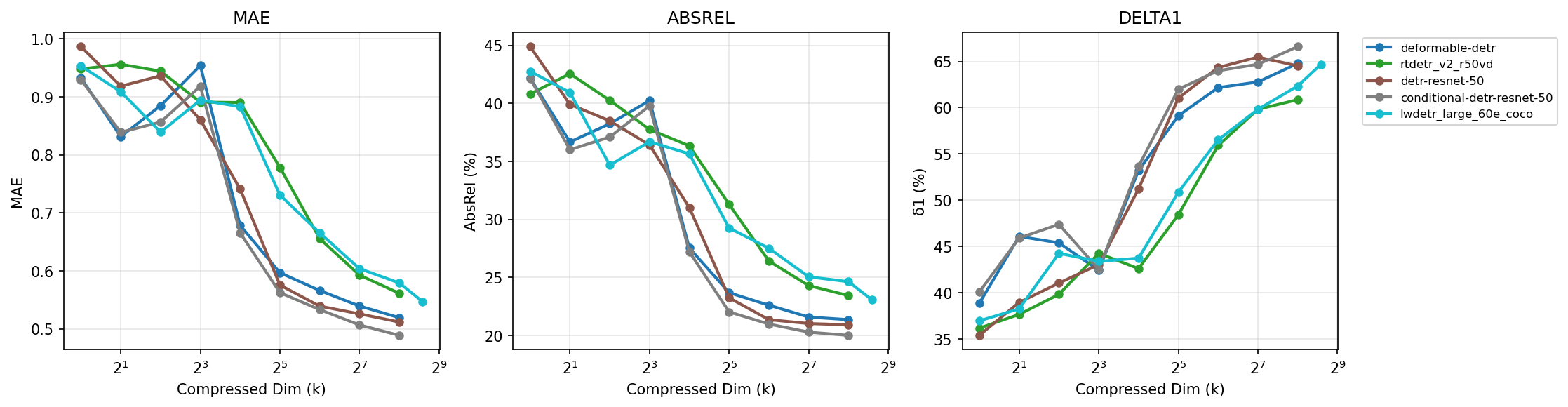}
\end{minipage}
\caption{Monocular depth estimation with MLP probes under PCA compression. Top: Virtual KITTI\,2. Bottom: NYUv2. We see rapid performance saturation as $k$ increases, indicating depth information is concentrated in a relatively low-dimensional subspace.}
\label{fig:compression-mlp}
\end{figure}

\textbf{Compressibility of Embeddings.} In this experiment, we investigate the extent to which object embeddings in the last decoder layer can be linearly compressed while retaining the ability to decode 3D information. We focus on the monocular depth estimation task. We first extract query embeddings, then compress them from $D$ to $k<D$ dimensions for values of $k$ from $2^0$ to the original representation size using PCA. Figure~\ref{fig:compression-mlp} shows the results. Across all DETR family models, performance improves rapidly as the PCA dimension increases, and then saturates beyond a moderate dimensionality. This trend suggests that depth-relevant information is concentrated in a relatively low-dimensional subspace of the embedding space. We also observe that real-time variants tend to require more retained dimensions to reach comparable performance, which is consistent with earlier observations regarding the under-performance of the representations produced by these models.\\

\begin{table}[t]
\centering
\caption{Monocular depth estimation with ablation on probe input and probe capacity. We compare embedding probes vs. bounding box  probes on the same Virtual KITTI\,2/NYUv2 experimental protocol.}
\label{tab:depth-detr-ablation}
\small
\setlength{\tabcolsep}{4pt}
\resizebox{\linewidth}{!}{%
\begin{tabular}{l ccc ccc}
\toprule
 & \multicolumn{3}{c}{\textbf{Virtual KITTI\,2}} & \multicolumn{3}{c}{\textbf{NYUv2}} \\
\cmidrule(lr){2-4} \cmidrule(lr){5-7}
\textbf{Method} & MAE\,$\downarrow$ & AbsRel (\%)\,$\downarrow$ & $\delta_1$\,$\uparrow$ & MAE\,$\downarrow$ & AbsRel (\%)\,$\downarrow$ & $\delta_1$\,$\uparrow$ \\
\midrule
DETR Bbox probe (Linear) & 4.27 & 32.67 & 44.69 & 0.87 & 37.61 & 41.74 \\
DETR Embedding probe (Linear) & \textbf{1.01} & \textbf{8.29} & \textbf{93.10} & \textbf{0.58} & \textbf{24.30} & \textbf{57.22} \\
\midrule
DETR Bbox probe (MLP) & 1.02 & 7.09 & 95.76 & 0.77 & 32.98 & 51.83 \\
DETR Embedding probe (MLP) & \textbf{0.54} & \textbf{3.62} & \textbf{98.81} & \textbf{0.51} & \textbf{20.20} & \textbf{65.48} \\
\bottomrule
\end{tabular}%
}
\end{table}

\noindent\textbf{Latent Representation Ablation.} In this experiment, we examine the predictive lift of object embeddings compared to 2D bounding box representations. We construct probing datasets for the 2D bounding box case exactly as for query embeddings except that we substitute the predicted bounding box representation $\mbf{b}$ for the query embedding $\mbf{q}$. Table~\ref{tab:depth-detr-ablation} shows the results for both linear and MLP probes using the DETR model on the monocular depth estimation task. We provide results on both the Virtual KITTI\,2 and NYUv2 datasets. Across both linear and MLP cases, the latent embedding probes dramatically outperform bounding box probes. This consistent margin shows that query embeddings contain substantial depth-relevant information beyond what is decoded into the 2D bounding box representation, likely capturing additional cues such as semantic context, category priors, and other high-level signals that correlate with object depth.

\section{Conclusions}\label{sec:conclusions}

In this paper, we study the extent to which 2D DETR family models that are trained without explicit 3D supervision encode object-level 3D properties in their latent object embeddings. We establish methods for constructing probe datasets to assess the understanding of 3D object-level properties, and apply the methodology to the tasks of object-centric monocular depth estimation and 3D object localization. Our results on monocular depth estimation show that 2D DETR models learn representations that can out-perform a zero-shot application of Depth Anything 2 on some scene types. Further, our results on 3D object localization show that 2D DETR models learn representations that can be decoded into high-quality location estimates with fractions of a meter more error than MonoDETR, a model trained specifically for this task using full 3D bounding box supervision. Taken together, these results suggest that 2D DETR models learn latent representations that encode a surprising amount of 3D structure. 

Future work includes assessing performance with respect to additional 3D properties, delving deeper into the representation gap between the light weight and real time DETR model variants and the more standard model variants, and leveraging these insights to inform the development of novel 3D DETR model architectures.

\section*{Acknowledgements}\label{sec:acknowledgement}

This research was sponsored by the U.S. Army Research Laboratory and was accomplished under Cooperative Agreement Number W911NF-17-2-0196. The views and conclusions contained in this document are those of the authors and should not be interpreted as representing the official policies, either expressed or implied, of the Army Research Laboratory or the U.S. Government. The U.S. Government is authorized to reproduce and distribute reprints for Government purposes notwithstanding any copyright notation herein.
\bibliographystyle{splncs04}
\bibliography{refs}

@article{khan2022transformers,
  title={Transformers in vision: A survey},
  author={Khan, Salman and Naseer, Muzammal and Hayat, Munawar and Zamir, Syed Waqas and Khan, Fahad Shahbaz and Shah, Mubarak},
  journal={ACM Computing Surveys},
  volume={54},
  number={10s},
  pages={1--41},
  year={2022},
  publisher={ACM New York, NY}
}

@inproceedings{maskformer_NEURIPS2021,
 author = {Cheng, Bowen and Schwing, Alex and Kirillov, Alexander},
 booktitle = {Advances in Neural Information Processing Systems},
 editor = {M. Ranzato and A. Beygelzimer and Y. Dauphin and P.S. Liang and J. Wortman Vaughan},
 pages = {17864--17875},
 publisher = {Curran Associates, Inc.},
 title = {Per-Pixel Classification is Not All You Need for Semantic Segmentation},
 volume = {34},
 year = {2021}
}

@InProceedings{tokenpose_2021_ICCV,
    author    = {Li, Yanjie and Zhang, Shoukui and Wang, Zhicheng and Yang, Sen and Yang, Wankou and Xia, Shu-Tao and Zhou, Erjin},
    title     = {TokenPose: Learning Keypoint Tokens for Human Pose Estimation},
    booktitle = {Proceedings of the IEEE/CVF International Conference on Computer Vision},
    month     = {October},
    year      = {2021},
    pages     = {11313-11322}
}

@inproceedings{detr3d_crl22,
  title={Detr3d: 3d object detection from multi-view images via 3d-to-2d queries},
  author={Wang, Yue and Guizilini, Vitor Campagnolo and Zhang, Tianyuan and Wang, Yilun and Zhao, Hang and Solomon, Justin},
  booktitle={Conference on Robot Learning},
  pages={180--191},
  year={2022},
  organization={PMLR}
}

@inproceedings{redmon2016you,
  title={You only look once: Unified, real-time object detection},
  author={Redmon, Joseph and Divvala, Santosh and Girshick, Ross and Farhadi, Ali},
  booktitle={Proceedings of the IEEE Conference on Computer Vision and Pattern Recognition},
  pages={779--788},
  year={2016}
}

@inproceedings{carion2020end,
  title={End-to-end object detection with transformers},
  author={Carion, Nicolas and Massa, Francisco and Synnaeve, Gabriel and Usunier, Nicolas and Kirillov, Alexander and Zagoruyko, Sergey},
  booktitle={European Conference on Computer Vision},
  pages={213--229},
  year={2020},
  organization={Springer}
}

@article{zhu2020deformable,
  title={Deformable detr: Deformable transformers for end-to-end object detection},
  author={Zhu, Xizhou and Su, Weijie and Lu, Lewei and Li, Bin and Wang, Xiaogang and Dai, Jifeng},
  journal={arXiv preprint arXiv:2010.04159},
  year={2020}
}

@inproceedings{meng2021conditional,
  title={Conditional detr for fast training convergence},
  author={Meng, Depu and Chen, Xiaokang and Fan, Zejia and Zeng, Gang and Li, Houqiang and Yuan, Yuhui and Sun, Lei and Wang, Jingdong},
  booktitle={Proceedings of the IEEE/CVF International Conference on Computer Vision},
  pages={3651--3660},
  year={2021}
}

@inproceedings{zhao2024detrs,
  title={Detrs beat yolos on real-time object detection},
  author={Zhao, Yian and Lv, Wenyu and Xu, Shangliang and Wei, Jinman and Wang, Guanzhong and Dang, Qingqing and Liu, Yi and Chen, Jie},
  booktitle={Proceedings of the IEEE/CVF Conference on Computer Vision and Pattern Recognition},
  pages={16965--16974},
  year={2024}
}

@article{lv2024rt,
  title={Rt-detrv2: Improved baseline with bag-of-freebies for real-time detection transformer},
  author={Lv, Wenyu and Zhao, Yian and Chang, Qinyao and Huang, Kui and Wang, Guanzhong and Liu, Yi},
  journal={arXiv preprint arXiv:2407.17140},
  year={2024}
}

@article{chen2024lw,
  title={Lw-detr: A transformer replacement to yolo for real-time detection},
  author={Chen, Qiang and Su, Xiangbo and Zhang, Xinyu and Wang, Jian and Chen, Jiahui and Shen, Yunpeng and Han, Chuchu and Chen, Ziliang and Xu, Weixiang and Li, Fanrong and others},
  journal={arXiv preprint arXiv:2406.03459},
  year={2024}
}

@article{oquab2024dinov2,
  title={DINOv2: Learning Robust Visual Features without Supervision},
  author={Oquab, Maxime and Darcet, Timoth{\'e}e and Moutakanni, Th{\'e}o and Vo, Huy and Szafraniec, Marc and Khalidov, Vasil and Fernandez, Pierre and Haziza, Daniel and Massa, Francisco and El-Nouby, Alaaeldin and others},
  journal={Transactions on Machine Learning Research},
  year={2024}
}

@inproceedings{ranftl2021vision,
  title={Vision transformers for dense prediction},
  author={Ranftl, Ren{\'e} and Bochkovskiy, Alexey and Koltun, Vladlen},
  booktitle={Proceedings of the IEEE/CVF International Conference on Computer Vision},
  pages={12179--12188},
  year={2021}
}

@inproceedings{lin2014microsoft,
  title={Microsoft coco: Common objects in context},
  author={Lin, Tsung-Yi and Maire, Michael and Belongie, Serge and Hays, James and Perona, Pietro and Ramanan, Deva and Doll{\'a}r, Piotr and Zitnick, C Lawrence},
  booktitle={European Conference on Computer Vision},
  pages={740--755},
  year={2014},
  organization={Springer}
}

@inproceedings{he2016deep,
  title={Deep residual learning for image recognition},
  author={He, Kaiming and Zhang, Xiangyu and Ren, Shaoqing and Sun, Jian},
  booktitle={Proceedings of the IEEE Conference on Computer Vision and Pattern Recognition},
  pages={770--778},
  year={2016}
}

@article{yang2024depth,
  title={Depth anything v2},
  author={Yang, Lihe and Kang, Bingyi and Huang, Zilong and Zhao, Zhen and Xu, Xiaogang and Feng, Jiashi and Zhao, Hengshuang},
  journal={Advances in Neural Information Processing Systems},
  volume={37},
  pages={21875--21911},
  year={2024}
}

@inproceedings{silberman2012indoor,
  title={Indoor segmentation and support inference from rgbd images},
  author={Silberman, Nathan and Hoiem, Derek and Kohli, Pushmeet and Fergus, Rob},
  booktitle={European Conference on Computer Vision},
  pages={746--760},
  year={2012},
  organization={Springer}
}

@article{cabon2020virtual,
  title={Virtual kitti 2},
  author={Cabon, Yohann and Murray, Naila and Humenberger, Martin},
  journal={arXiv preprint arXiv:2001.10773},
  year={2020}
}

@inproceedings{wolf2020transformers,
  title={Transformers: State-of-the-art natural language processing},
  author={Wolf, Thomas and Debut, Lysandre and Sanh, Victor and Chaumond, Julien and Delangue, Clement and Moi, Anthony and Cistac, Pierric and Rault, Tim and Louf, R{\'e}mi and Funtowicz, Morgan and others},
  booktitle={Proceedings of the 2020 Conference on Empirical Methods in Natural Language Processing: System Demonstrations},
  pages={38--45},
  year={2020}
}

@inproceedings{zhang2023monodetr,
  title={MonoDETR: Depth-guided transformer for monocular 3d object detection},
  author={Zhang, Renrui and Qiu, Han and Wang, Tai and Guo, Ziyu and Cui, Ziteng and Qiao, Yu and Li, Hongsheng and Gao, Peng},
  booktitle={Proceedings of the IEEE/CVF International Conference on Computer Vision},
  pages={9155--9166},
  year={2023}
}

@inproceedings{el2024probing,
  title={Probing the 3d awareness of visual foundation models},
  author={El Banani, Mohamed and Raj, Amit and Maninis, Kevis-Kokitsi and Kar, Abhishek and Li, Yuanzhen and Rubinstein, Michael and Sun, Deqing and Guibas, Leonidas and Johnson, Justin and Jampani, Varun},
  booktitle={Proceedings of the IEEE/CVF Conference on Computer Vision and Pattern Recognition},
  pages={21795--21806},
  year={2024}
}

@article{zhan2024general,
  title={A general protocol to probe large vision models for 3d physical understanding},
  author={Zhan, Guanqi and Zheng, Chuanxia and Xie, Weidi and Zisserman, Andrew},
  journal={Advances in Neural Information Processing Systems},
  volume={37},
  pages={43468--43498},
  year={2024}
}

@article{vaswani2017attention,
  title={Attention is all you need},
  author={Vaswani, Ashish and Shazeer, Noam and Parmar, Niki and Uszkoreit, Jakob and Jones, Llion and Gomez, Aidan N and Kaiser, {\L}ukasz and Polosukhin, Illia},
  journal={Advances in Neural Information Processing Systems},
  volume={30},
  year={2017}
}

@inproceedings{geiger2012we,
  title={Are we ready for autonomous driving? the kitti vision benchmark suite},
  author={Geiger, Andreas and Lenz, Philip and Urtasun, Raquel},
  booktitle={2012 IEEE Conference on Computer Vision and Pattern Recognition},
  pages={3354--3361},
  year={2012},
  organization={IEEE}
}

@article{alain2016understanding,
  title={Understanding intermediate layers using linear classifier probes},
  author={Alain, Guillaume and Bengio, Yoshua},
  journal={arXiv preprint arXiv:1610.01644},
  year={2016}
}

@inproceedings{wagner2024forgetting,
  title={Forgetting analysis by module probing for online object detection with faster R-CNN},
  author={Wagner, Baptiste and Pellerin, Denis and Huet, Sylvain},
  booktitle={2024 32nd European Signal Processing Conference},
  pages={576--580},
  year={2024},
  organization={IEEE}
}

@inproceedings{mikriukov2023evaluating,
  title={Evaluating the stability of semantic concept representations in CNNs for robust explainability},
  author={Mikriukov, Georgii and Schwalbe, Gesina and Hellert, Christian and Bade, Korinna},
  booktitle={World Conference on Explainable Artificial Intelligence},
  pages={499--524},
  year={2023},
  organization={Springer}
}

@article{marks2023geometry,
  title={The geometry of truth: Emergent linear structure in large language model representations of true/false datasets},
  author={Marks, Samuel and Tegmark, Max},
  journal={arXiv preprint arXiv:2310.06824},
  year={2023}
}

@inproceedings{azaria2023internal,
  title={The internal state of an LLM knows when it’s lying},
  author={Azaria, Amos and Mitchell, Tom},
  booktitle={Findings of the Association for Computational Linguistics: Conference on Empirical Methods in Natural Language Processing 2023},
  pages={967--976},
  year={2023}
}

@inproceedings{park2024linear,
  title={The linear representation hypothesis and the geometry of large language models},
  author={Park, Kiho and Choe, Yo Joong and Veitch, Victor},
  booktitle={Proceedings of the 41st International Conference on Machine Learning},
  pages={39643--39666},
  year={2024}
}

@inproceedings{danier2025depthcues,
  title={DepthCues: Evaluating monocular depth perception in large vision models},
  author={Danier, Duolikun and Ayg{\"u}n, Mehmet and Li, Changjian and Bilen, Hakan and Mac Aodha, Oisin},
  booktitle={Proceedings of the Computer Vision and Pattern Recognition Conference},
  pages={20049--20059},
  year={2025}
}

@inproceedings{gairola2025probe,
  title={How to probe: Simple yet effective techniques for improving post-hoc explanations},
  author={Gairola, Siddhartha and B{\"o}hle, Moritz and Locatello, Francesco and Schiele, Bernt},
  booktitle={International Conference on Learning Representations},
  year={2025}
}

@inproceedings{li2023evaluating,
  title={Evaluating representations with readout model switching},
  author={Li, Yazhe and Bornschein, Jorg and Hutter, Marcus},
  booktitle={International Conference on Learning Representations},
  year={2023}
}

@inproceedings{caron2021emerging,
  title={Emerging properties in self-supervised vision transformers},
  author={Caron, Mathilde and Touvron, Hugo and Misra, Ishan and J{\'e}gou, Herv{\'e} and Mairal, Julien and Bojanowski, Piotr and Joulin, Armand},
  booktitle={Proceedings of the IEEE/CVF International Conference on Computer Vision},
  pages={9650--9660},
  year={2021}
}

@inproceedings{rombach2022high,
  title={High-resolution image synthesis with latent diffusion models},
  author={Rombach, Robin and Blattmann, Andreas and Lorenz, Dominik and Esser, Patrick and Ommer, Bj{\"o}rn},
  booktitle={Proceedings of the IEEE/CVF Conference on Computer Vision and Pattern Recognition},
  pages={10684--10695},
  year={2022}
}

@article{belinkov2022probing,
  title={Probing classifiers: Promises, shortcomings, and advances},
  author={Belinkov, Yonatan},
  journal={Computational Linguistics},
  volume={48},
  number={1},
  pages={207--219},
  year={2022}
}
\appendix
\clearpage
\section{Implementation Details}
\label{app:impl}

\textbf{DETR families and baselines checkpoints.}
All 2D detector checkpoints are loaded from the HuggingFace model hub using model-specific processor and detector classes. The exact checkpoints used in our experiments are listed in Table~\ref{tab:appendix-checkpoints}. For the dense depth baselines, we use the metric Depth Anything v2 checkpoints specialized for outdoor and indoor scenes. For the 3D detection baseline, we use the official MonoDETR repository and released best checkpoint from authors.

\begin{table}[t]
\centering
\caption{Exact checkpoints used in our experiments.}
\label{tab:appendix-checkpoints}
\scriptsize
\setlength{\tabcolsep}{4pt}
\begin{tabularx}{\linewidth}{@{}p{0.29\linewidth}X@{}}
\toprule
\textbf{Model} & \textbf{Checkpoint URL} \\
\midrule
DETR & \url{https://huggingface.co/facebook/detr-resnet-50} \\
Conditional DETR & \url{https://huggingface.co/microsoft/conditional-detr-resnet-50} \\
Deformable DETR & \url{https://huggingface.co/SenseTime/deformable-detr} \\
RT-DETR v2 & \url{https://huggingface.co/PekingU/rtdetr_v2_r50vd} \\
LW-DETR & \url{https://huggingface.co/AnnaZhang/lwdetr_large_60e_coco} \\
\midrule
Depth Anything 2 (outdoor) & \url{https://huggingface.co/depth-anything/Depth-Anything-V2-Metric-Outdoor-Base-hf} \\
Depth Anything 2 (indoor) & \url{https://huggingface.co/depth-anything/Depth-Anything-V2-Metric-Indoor-Base-hf} \\
MonoDETR repository & \url{https://github.com/ZrrSkywalker/MonoDETR} \\
MonoDETR checkpoint & \url{https://drive.google.com/file/d/1d8fbAt-CQF-IN8UEHuw3NimmfONhH6iA/view?usp=sharing} \\
\bottomrule
\end{tabularx}
\end{table}

\textbf{Probe architectures.}
For every experiment, the detector weights are frozen and only the probe is trained. The linear probe consists of a single affine map from the detector embedding to the target space. The non-linear probe is a two-layer MLP with ReLU activations and hidden dimension $256$. Unless otherwise noted, we use a batch size of $512$, learning rate $10^{-3}$, and mean squared error as the optimization objective for regression tasks. For the MLP depth experiments, we use $1000$ epochs with a $20$-epoch warmup followed by cosine decay. All detector query embeddings have dimension $256$, except for LW-DETR whose decoder embeddings have dimension $384$.

\section{Depth Probe Data Construction}
\label{app:depth-data}

\subsection{Virtual KITTI 2}

For the Virtual KITTI 2 experiments, we use images from \texttt{Camera\_0}. Each detector is run on every image, detections are filtered using an objectness threshold of $\tau=0.5$, and only detections mapped to the \texttt{car} class in the detector label space are retained. Ground-truth depth supervision is obtained by sampling the Virtual KITTI 2 metric depth map at the center of the anchor box. We remove detections whose sampled depth lies outside the range $[0.5, 100]$ meters.

\subsection{NYU Depth v2}

For NYUv2, we run the same extraction procedure on RGB images from the labeled RGB-D split. Unlike Virtual KITTI 2, we keep detections from all foreground classes. The objectness threshold is again fixed at $\tau=0.5$. Ground-truth depth is sampled at the predicted box center, and detections with sampled depth outside the range $[0.1, 10]$ meters are discarded. The resulting aligned NYUv2 depth data set contains $5750$ object instances, which are split into $4600$ training instances and $1150$ test instances.

\subsection{Depth Baselines}

For Depth Anything, we run the dense depth model once per image and sample the predicted depth at the same 2D box centers used to supervise the probe. This yields a detector-aligned object-centric depth baseline that is directly comparable to the probe outputs. For MonoDETR, we evaluate two variants in the depth experiment. In the zero-shot setting, we match MonoDETR predictions to the aligned 2D detections using 2D IoU and directly compare the predicted object-center depth against the target depth used for the probe task. In the learned-probe setting, we extract MonoDETR decoder query embeddings, align them to the same detected objects, and then train an MLP probe.

\section{Additional KITTI 3D Center Prediction Details}
\label{app:kitti-details}

For the 3D object location experiment,  we run each frozen 2D DETR detector on the KITTI images, retain detections assigned to the vehicle category of interest, and match detector boxes to KITTI annotations by greedy one-to-one IoU matching with threshold $0.5$. Unmatched detections are discarded. For each matched detection, the probe target is the KITTI 3D box center represented in camera coordinates. MonoDETR is evaluated on the same validation split. Its predictions are filtered by confidence score and then matched to KITTI ground truth using the same IoU-based rule. This ensures that the DETR-family probes and the MonoDETR baseline are evaluated under the same object-level matching criterion.

\section{Qualitative Depth Examples}
\label{app:qual-depth}

\begin{figure*}[t]
\centering
\includegraphics[width=\linewidth]{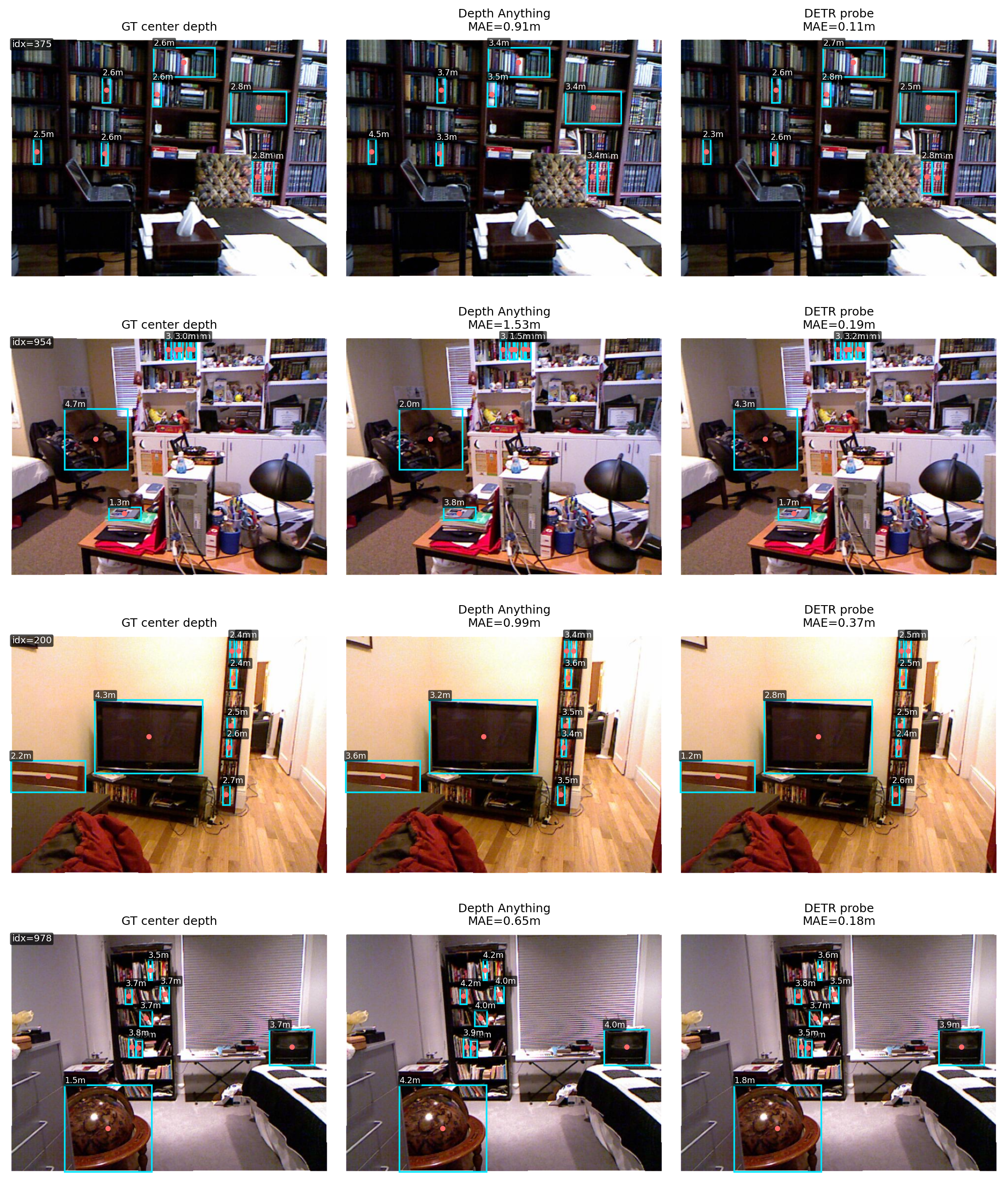}
\caption{Qualitative object-centric depth predictions on NYUv2. From left to right: input image with predicted boxes, centers, and ground-truth center depth; Depth Anything 2 center depth; and DETR probe center depth.}
\label{fig:appendix-qual-nyuv2}
\end{figure*}

\begin{figure*}[t]
\centering
\includegraphics[width=\linewidth]{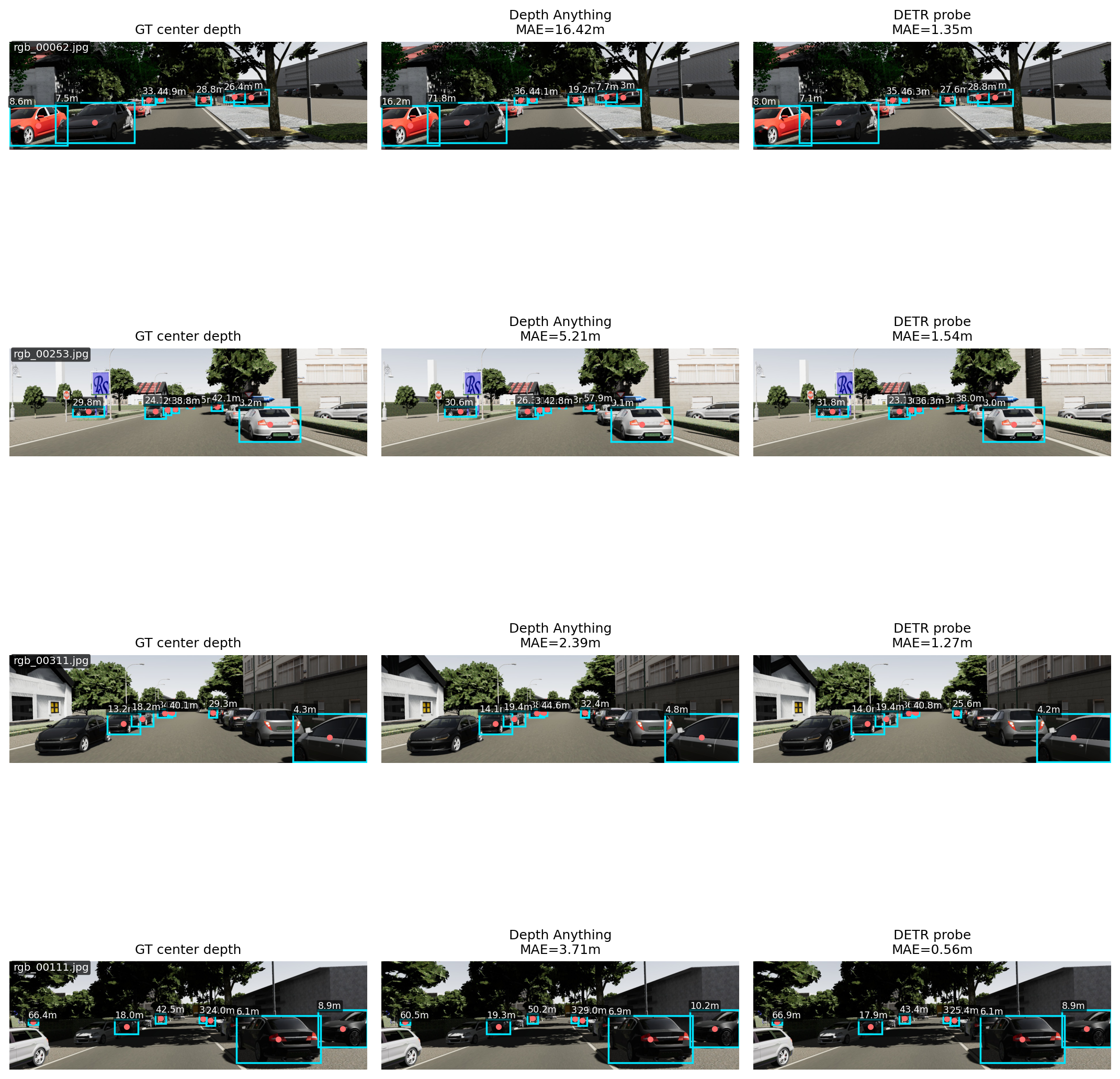}
\caption{Qualitative object-centric depth predictions on Virtual KITTI\,2. From left to right: input image with predicted boxes, centers, and ground-truth center depth; Depth Anything 2 center depth; and DETR probe center depth.}
\label{fig:appendix-qual-vkitti2}
\end{figure*}

Figure~\ref{fig:appendix-qual-nyuv2} and Figure~\ref{fig:appendix-qual-vkitti2} provide qualitative comparisons between ground-truth object-center depth, Depth Anything 2, and the DETR MLP probe. Each row corresponds to one image from the test split.

\end{document}